\documentclass[runningheads]{llncs}
\usepackage{cite}
\usepackage{times}
\usepackage{epsfig}
\usepackage{graphicx}
\usepackage{amsmath}
\usepackage{amssymb}
\usepackage{float}
\usepackage{booktabs}
\usepackage{caption}
\usepackage{subcaption}
\usepackage[nolist]{acronym}
\graphicspath{{figs/}}

\begin{document}
\title{Towards Comparable Knowledge Distillation in Semantic Image Segmentation}
%
%
\author{Onno Niemann\inst{1}\orcidID{0009-0004-6131-529X} \and
Christopher Vox\inst{2}\orcidID{0009-0003-4050-5070} \and
Thorben Werner\inst{1}\orcidID{0000-0002-5944-6045}}
%
%
\institute{University of Hildesheim, 31141 Hildesheim, Germany
\and
Volkswagen AG, Berliner Ring 2, 38440 Wolfsburg, Germany
\email{onno.niemann@uni-hildesheim.de}}
%
\maketitle              
\begin{abstract}
Knowledge Distillation (KD) is one proposed solution to large model sizes and slow inference speed in semantic segmentation. In our research we identify 25 proposed distillation loss terms from 14 publications in the last 4 years. Unfortunately, a comparison of terms based on published results is often impossible, because of differences in training configurations. A good illustration of this problem is the comparison of two publications from 2022. 
Using the same models and dataset, Structural and Statistical Texture Distillation (SSTKD) \cite{structural} 
reports an increase of student mIoU of 4.54 and a final performance of 29.19, while Adaptive Perspective Distillation (APD) \cite{perspective} only improves student performance by 2.06 percentage points, but achieves a final performance of 39.25. 
The reason for such extreme differences is often a suboptimal choice of hyperparameters and a resulting underperformance of the student model used as reference point.

In our work, we reveal problems of insufficient hyperparameter tuning by showing that distillation improvements of two widely accepted frameworks, SKD and IFVD, vanish when hyperparameters are optimized sufficiently. To improve comparability of future research in the field, we establish a solid baseline for three datasets and two student models and provide extensive information on hyperparameter tuning. We find that only two out of eight techniques can compete with our simple baseline on the ADE20K dataset.

\end{abstract}

\begin{acronym}
\acro{miou}[mIoU]{Mean Intersection over Union}
\end{acronym}

\section{Introduction}

Advances in Deep Learning techniques for Semantic Image Segmentation brought major performance improvements to fields such as autonomous driving, medical image analysis, robotic perception and video surveillance. As these performance gains often came at the price of increased model complexity and required computational power, efficient deep learning techniques have become increasingly relevant \cite{Menghani}.

Two techniques, which start with a large model and make it more efficient are model Pruning and Quantization. 
Pruning shrinks a model by dropping less important nodes and Quantization reduces the numerical precision of weights. Knowledge distillation takes a different approach and does not change model efficiency during training. Instead it starts with a small model (student) and improves its performance by leveraging guidance from a more complex model (teacher). Teacher model weights are frozen and knowledge is distilled into the student by making an addition to the student loss, which penalizes differences in student and teacher output. Incentivizing the student to mimic the more complex behaviour of the teacher can significantly lift its performance.

KD was introduced in the image classification domain and originally the distillation loss was applied to student and teacher only at output-level. The extension of teacher guidance to intermediate layers and the transfer to semantic segmentation have been studied in various publications. Re-using the most basic, output-level distillation loss in a segmentation context is straightforward, as it can be applied on pixel instead of image-level in the student and teacher output. 
However, several papers criticize that this naive approach treats pixels in isolation and introduce more complex techniques. Most of them build upon the naive pixel-wise distillation by adding loss terms to it. A problem with these published methods is that a significant share focuses solely on performance improvement in their own training framework and fails to provide a good baseline for comparison against other literature.

The dimension of this comparability problem is illustrated by the overview over important publications and their results we provide in Table \ref{tab:over}. The comparison clearly shows that the baseline performance of the student model varies strongly between publications, making it hard to compare the quality of the proposed techniques based on the lift they provide. To address this problem of comparability, we perform an extensive hyperparameter optimization of the most fundamental of KD frameworks for semantic segmentation and provide detailed information on optimal hyperparameters for training of two student models on three datasets. As a byproduct of this optimization, we find that the temperature parameter used to "soften" student and teacher output in image classification can improve distillation in segmentation, although many publications in the field ignore it.

In summary, we point out challenges in comparing different methods and present a training procedure, which sets the ground for a fair comparison of KD techniques. We further put the performance of three commonly used loss terms into perspective by comparing them to our achieved results.  

\section{Related Work}\label{chap:rel}

\paragraph{Semantic Segmentation.} Most earlier image segmentation techniques were based on Partial Differential Equations or Random Forest methods \cite{brief}, before the advent of deep learning sparked a series of publications leveraging the powers of Convolutional Neural Networks (CNNs). The majority of this early CNN-based segmentation research focusses on improving model performance, which is achieved by adding skip connections \cite{fcn} and decoder networks \cite{deconv, unet, segnet}, or concatenating convolutions at different scales in the pyramid pooling module of Pyramid Scence Parsing Networks (PSPNets) \cite{pspnet}. More recent publications investigate the application of transformer based models \cite{rethinking, segmenter, segformer}. Another branch of segmentation research focuses more on model efficiency instead of performance. Real-time semantic segmentation aims for fast inference speed while maintaining performance \cite{icnet, enet, dfanet, bisenet}, but there is always a trade-off between performance and efficiency.

\paragraph{Knowledge Distillation in Classification.} KD aims to increase the performance of a compact student model by leveraging guidance from a large teacher model during training. It is one of the most widely used model compression techniques due to its broad applicability \cite{pre-train}. Unlike Model Pruning or Quantization, KD is model agnostic \cite{patient} as no requirements are imposed on the teacher or student model. Additionally, KD allows leveraging unlabelled data as the teacher can provide soft labels for student training \cite{pre-train}. 

Generally, the teacher model weights are frozen during student training and the student is encouraged to mimic the teacher's output logits by training on a weighted combination of standard cross entropy and KD loss \cite{caruana, Distilling} (Eq. \ref{eq:loss}). 
The cross entropy term (Eq. \ref{eq:ce}) ensures performance on the labeled training data while the distillation term (Eq. \ref{eq:kd}) penalizes deviations from the teacher output. The weight of the distillation loss, $\lambda$, is a hyperparameter of student training. Scaling output logits of both models by $\tau > 1$ is crucial for a successfull distillation of knowledge in image classification \cite{Distilling} and extending the level of student teacher matching to intermediate feature layers can give a further boost \cite{fitnets, attention}.


\begin{equation}\label{eq:loss}
L_S=L_{CE}+\lambda * L_{KD}
\end{equation}

\begin{equation}\label{eq:ce}
L_{CE}(z_s, y)=-\sum_{c=1}^Cy_c\log \sigma_c(z_s)
\end{equation}

\begin{equation}\label{eq:kd}
    L_{KD}(z_s, z_t)=-\tau^2\sum_{c=1}^C \sigma_c(\frac{z_t}{\tau})\log \sigma_c(\frac{z_s}{\tau})
\end{equation} 

Despite the simplicity of the classical KD framework, the underlying mechanisms are still not well-understood. 
Commonly, the success of KD is associated with the information contained in teacher predictions for wrong classes, referred to as "dark knowledge" \cite{smoothing}. However, phenomena such as students outperforming teachers when trained solely on teacher output and high disagreement between student and teacher predictions suggest that there are other explanations for the success of KD \cite{work?}. 
Two alternative explanations are that teacher guidance has a regularizing effect similar to label smoothing \cite{smoothing} and that teacher output provides a sample-wise importance weighting, making the student focus on samples of low teacher confidence \cite{bornagain}.

\paragraph{KD in Image Segmentation.}
Applying the classical KD framework to semantic image segmentation problems is straight-forward. The distillation loss in  Equation \ref{eq:kd} is usually applied on image-level, summing over all classes, but since image segmentation is essentially pixel-wise classification, it can be applied to each pixel in the image instead. This most basic distillation loss is referred to as pixel-wise distillation ($L_{PI}$) and is a useful baseline for more complex distillation schemes. Since pixel-wise distillation treats each pixel in isolation and ignores the fact that segmentation depends strongly on contextual information, various alternative distillation techniques exist, many of which use pixel-wise distillation as one part of their framework.

The earliest proposed technique investigates a "consistency" loss comparing regional differences in student and teacher output \cite{consistency} aiming for a more contextual distillation of knowledge by matching the distance of the center pixel and an 8-neighborhood. Instead of a direct matching of student and teacher output, in Knowledge Adaptation (KA) \cite{adaptation} teacher output is compressed to a more dense latent space by an autoencoder before being compared to student logits. Additionally, an affinity loss term is introduced to better capture long-range dependencies.

Structured Knowledge Distillation (SKD) \cite{Structured} is the most cited publication in the field and uses a combination of three loss terms, one of them the basic pixel-wise loss. Again the other two loss terms are supposed to focus more on contextual information in both intermediate and output layers. The pair-wise loss ($L_{PA}$) is based on a pair-wise Markov random field framework and encourages students to mimic the teacher at intermediate layers. The holistic loss ($L_{HO}$) requires an additional discriminator model with the task to differentiate student and teacher output. Student and discriminator compete with each other following a similar training protocol as used to train Generative Adversarial Networks\cite{gan}, ultimately leading to the student output being as similar as possible to the teacher output \cite{Structured}. 

Building upon SKD, Intra-Class Feature Variation Distillation (IFVD) \cite{intra-class} also uses the pixel-wise and holistic loss terms, but replaces the pair-wise loss with the new IFV loss, $L_{IFV}$.
Per-class prototypes are calculated and the distance of intermediate features to the respective prototype is aligned between student and teacher. 
Another suggested approach, CSCACE \cite{csc}, introduces a Channel and Spatial Correlation (CSC) and an Adaptive Cross Entropy (ACE) term. CSC calculates correlation matrices between intermediate features and can be understood as an extension of the pair-wise loss of SKD. ACE combines the classic pixel-wise KD loss with the ground truth label by only using the teacher output when the teacher prediction is correct.

As one of the few approaches that modify the pixel-wise distillation loss, Channel-Wise Distillation (CWD) \cite{channel-wise} proposes a channel/class-wise normalization of student and teacher outputs before calculating the KL-divergence, a technique that is copied by several later methods.
%
Double Similarity Distillation (DSD) \cite{double} introduces a pixel-wise similarity loss, which matches intermediate student and teacher features by utilizing self-attention maps across multiple layers. 
A category-wise similarity distillation loss makes the student mimic the teacher at output level by minimizing the L2 distance between correlation matrices of student and teacher output.
%
Masked Generative Distillation (MGD) \cite{masked} is a more general technique, which can also be used for knowledge distillation in image classification or object detection. 
The proposed method masks some parts of the input while still requiring the student to mimic the full teacher output.

Inter-Class Distance Distillation (IDD) \cite{inter-class} and Feature-Augmented Knowledge Distillation (FAKD) \cite{feat-augment} use the pixel-wise-distillation loss with channel-wise normalization as suggested in CWD \cite{channel-wise}. IDD additionally includes all terms introduced by SKD and an inter-class feature distance loss following a similar reasoning as IFVD. 
FAKD does not introduce additional loss terms but proposes the perturbation of intermediate student features in multiple ways and training the student to mimic the teacher despite the applied perturbations. 

Structural and Statistical Texture Knowledge Distillation (SSTKD) \cite{structural} and Adaptive Perspective Distillation (APD) \cite{perspective} use the pixel-wise distillation loss.
SSTKD additionally includes the holistic loss of SKD and two novel loss terms, which encourage the student to mimic low-level texture information of the teacher \cite{structural}. 
%
The authors of APD argue that segmentation networks learn to generalize and thus acquire a universal perception. 
Their introduced adaptive perspective includes calculating class-wise average features of individual images. 
According to the authors, this process distills contextual information more explicitly. 
APD is the only reviewed technique that updates part of the teacher model during student training.

Self-attention and Self-distillation \cite{self-distill} introduces a self-attention loss term to make the student learn contextual information from the teacher and a layer-wise context distillation loss. The second term is different from other discussed techniques, as it is applied across student layers to ensure a consistent representation of contextual information in shallow layers.
Cross-Image Relational Knowledge Distillation (CIRKD) \cite{cirkd} again uses the pixel-wise distillation loss and uses three more loss terms. Unique about this method is the introduction of a pixel queue to make student models mimic the teachers distance to output for pixels of the same class from previous images. 
\section{Related Work Comparison}
A comparison of the discussed knowledge distillation frameworks is rarely straightforward.
A very common problem is differences in student and teacher architectures. Especially earlier methods use a variety of models and results cannot be compared to recent literature. 
Table \ref{tab:over} presents the results of different distillation frameworks that report performance on the most common choice of student and teacher model, PSPNets with ResNet101 and ResNet18 backbone, respectively. This choice of models has been the standard in the field since the publication of SKD \cite{Structured}, but Consistency \cite{consistency}, CSCACE \cite{csc}, KA \cite{adaptation} and SALC \cite{self-distill} use other architectures and are excluded from our table. 

\begin{table}
    \centering
    \caption{Comparison of reported student \ac{miou} of various techniques following the most common architectural setup.
    The "Ls" column  refers to the number of loss terms in addition to the CE loss. 
    "S only" shows the performance of the student model trained
    without teacher. 
    "+T "contains the best performance reported in the publication applying 
    all proposed loss terms. 
    Methods followed by * use the $L_{PI}$, the ones followed by $\dag$ additionally do channel-wise normalization before applying $L_{PI}$.}
    \label{tab:over}
    \begin{tabular}{lccccccc}\toprule
        & & \multicolumn{2}{c}{ADE20K} & \multicolumn{2}{c}{Cityscapes}
        \\\cmidrule(lr){3-4}\cmidrule(lr){5-6}
        & Ls & S only & + T & S only & + T 
        \\\midrule
        SKD\cite{Structured}* & 3 & 33.82 & 36.55 & 69.10 & 72.67\\
        IFVD\cite{intra-class}* & 3 & - & - & 69.10 & 74.54 \\
        CWD\cite{channel-wise}$\dag$ & 1 & 24.65 & 26.80 & 70.09 & 75.90\\
        DSD\cite{double} & 2 & 33.80 & 38.00 & 69.42 & 73.20\\
        MGD\cite{masked} & 1 & - & - & 69.85 & 74.10\\
        IDD\cite{inter-class}$\dag$ & 6 & 24.65 & 27.69 & 70.09 & 77.59\\
        FAKD\cite{feat-augment}$\dag$ & 1 & 29.42 & 35.30 & 68.99 & 74.75\\
        SSTKD\cite{structural}* & 4 & 24.65 & 29.19 & 69.10 & 75.15\\
        APD\cite{perspective}* & 3 & 37.19 & 39.25 & 74.15 & 75.68\\
        CIRKD\cite{cirkd} & 4 & - & - & 72.55 & 74.73 \\
        \bottomrule 
    \end{tabular}
\end{table}

Even though the techniques in Table \ref{tab:over} use the same model architectures we can observe significant performance differences for both the student-only training ("S only") and the full KD framework ("+T").
While the differences in "+T" are to be expected for different KD algorithms, the inconsistent baseline performance of the student model makes it very hard to compare different algorithms.

Additionally, the listed techniques often propose a number of loss terms ("Ls"), but do not provide comprehensive ablation studies that isolate the effect of individual terms across different datasets.
Finally, many results in recent literature are reported without standard deviations, with SKD being the only exemption in Table \ref{tab:over}.
This introduces another complicating factor when comparing results of KD algorithms.

For a specific example we look at APD vs. DSD:
Judging only by the final performance of 39.25, APD outperforms all other techniques on ADE20K. Considering its starting point of 37.19, however, student performance is only lifted by 2.06. As DSD achieves a lift of 4.20 using one loss term less, but only achieves a final performance of 38.00, it is unclear, which of the two techniques is superior.
\section{Methodology}\label{sec:methodology}
In this work, we present a series of experiments highlighting the importance of individual hyperparameter tuning for every combination of datasets and models.
We grid search initial learning rate $\mu(0)$ and regularization rate $\gamma$ for student with and without teacher model separately to ensure the best performance possible in both settings.
In this way we guarantee a fair performance comparison when measuring the lift from KD later. 
In a second step, we optimize the temperature parameter $\tau$ of the combined student and teacher training. 

We also fine-tune the weight of the pixel-wise distillation loss, $\lambda_{PI}$, and find an optimal performance for $\lambda_{PI}=1e-1$ for all experiments. 
Finally, we test different proposed loss terms in isolation to measure their effect on student performance by optimizing their respective loss weights.
\section{Experiments}
\subsection{Datasets and Evaluation Metric}
We evaluate our approach on three public semantic segmentation benchmark datasets, PascalVOC\cite{voc}, Cityscapes\cite{cityscapes}, and ADE20K\cite{ade}. All datasets have pixel-wise annotations.

\textbf{Cityscapes} is an urban street scene understanding dataset showing street scenes recorded in 50 cities. It contains 5,000 finely annotated images with labels from 19 classes. All images have dimension 2048x1024 and train, validation and test set contain 2,975, 500 and 1,525 images, respectively.

\textbf{ADE20K} is a complex scene understanding dataset containing 20K/2K/3K images of different sizes in train, validation and test set. Images show  objects, parts of objects and stuff from varying context and pixels are assigned one of 150 object and stuff class labels.

\textbf{PascalVOC} contains 1,464 images for training, 1,449 images for validation and a private test set. It has 20 different object classes and images are of varying size.

\textbf{Mean Intersection over Union} is the metric used in all our experiments. 

\subsection{Implementation Details}\label{sec:ImplementationDetails}
As most other relevant methods (Table \ref{tab:over}), we follow the architectural setup and training procedure of SKD \cite{Structured}. To be precise, this means using PSPNets \cite{pspnet} with different backbones as student and teacher models. The teacher model has a ResNet101 \cite{he2016deep} backbone and for the student we test a ResNet18\cite{he2016deep} and an EfficientNet-B0 \cite{tan2019efficientnet} backbone. For simplicity we will refer to the two student models as ResNet and EffNet students. 
The learning rate $\mu(i)$ decays over training steps according to Eq. \ref{eq:lr_scheduler}. 
\begin{equation} \label{eq:lr_scheduler}
   \mu (i)=\mu(0) * ( 1 - \frac{i}{\eta})^{0.9}\hspace{0.5mm}; \hspace{2mm} i := [1 \ldots \eta] 
\end{equation}
$\eta$ is the total number of training batches and i is the current batch. 
Unless otherwise stated the ResNet and EffNet student backbones are initialized with weights pre-trained on ImageNet. The pre-trained student weights were obtained from the torchvision package (v0.12) of the PyTorch library \cite{pytorch2019} for both backbones and the teacher weights were taken from \cite{Structured} for Cityscapes and from \cite{semseg2019} for the other datasets.

All experiments are conducted with a batch size of 8 and students are trained with crops of size $512\times 512$ on Cityscapes and $473\times 473$ on ADE20K and PascalVOC.

All experimental results are calculated on the validation datasets of Cityscapes, PascalVOC and ADE20K.
\subsection{The Impact of Temperature}\label{sec:tuning_tau}
The authors of the original KD framework \cite{Distilling} strongly emphasize the importance of logit scaling by a temperature parameter $\tau>1$ in image classification, 
the main reason for its success being the "softening" of teacher output class distributions \cite{Distilling}.
To show that teacher output distributions are "hard" also in image segmentation we analyze the effect of different values of $\tau$ on the Shannon Entropy \cite{shannon} of teacher output. 
%
When all probability mass is assigned to one class the Shannon Entropy is zero and when the probability mass is distributed evenly over all classes Shannon Entropy is maximal.

\begin{figure}[h]
\includegraphics[width=0.7\textwidth]{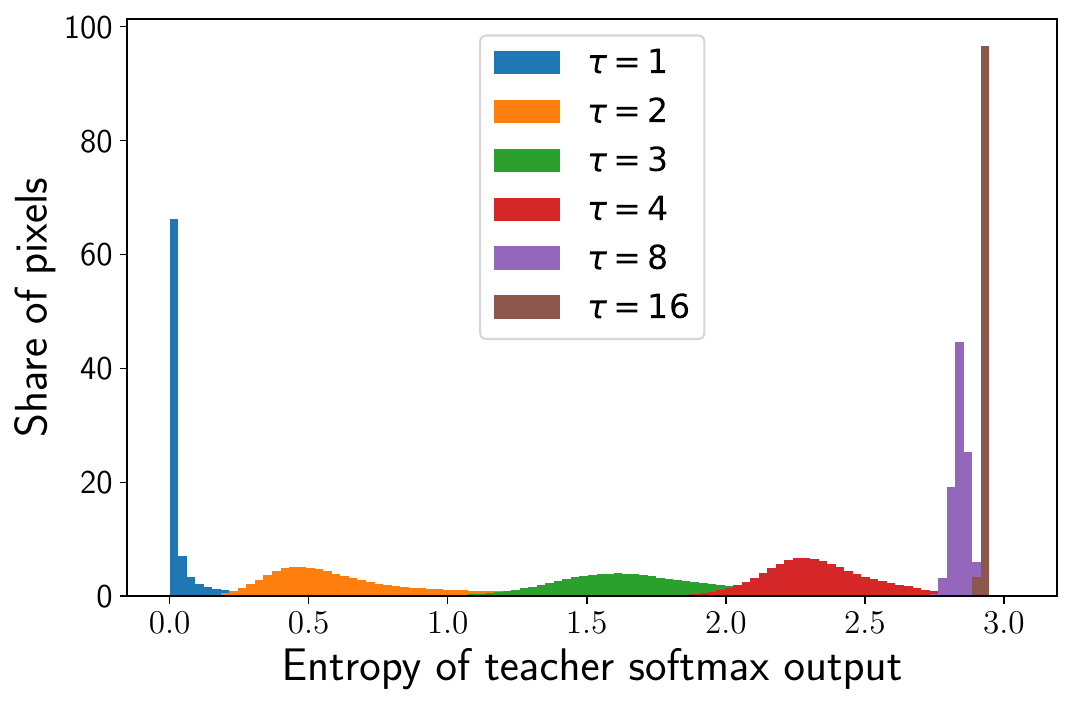}
\centering
\caption{Effect of logit scaling on Shannon Entropy of teacher output probability distributions over classes. 
The histograms are based on class probability distributions of 
randomly selected pixels from Cityscapes. 
Without scaling ($\tau=1$) over 60\% of pixels are assigned one class with a probability close to 1.}
\label{fig:entropy}
\end{figure}

We generate teacher output for 800 randomly selected images
from the Cityscapes dataset, resulting in probability distributions over the 19 classes for 209,715,200\;pixels. These distributions are scaled with different temperature values according to Equation \ref{eq:kd}.
The shares of Shannon Entropies over all pixels for different temperatures are shown in Figure\;\ref{fig:entropy}.
%
%
Temperatures of 1, 2, 4, 8, and 16 were chosen as \cite{Distilling} report successful distillation for values up to 20. 
An important observation is that when the teacher output is not scaled ($\tau=1$), the entropy strongly spikes at 0. More than 60\% of pixels are classified with a confidence close to 1, suggesting that teacher output might be too "hard" for efficient distillation of knowledge and higher temperatures might help. On the other hand, for a value of $\tau=16$ almost 100\% of distributions have an entropy of approximately 2.9, indicating an almost even distribution of probability mass over all classes. 
%
\subsection{Hyperparameter Optimization}
As described previously, we optimize hyperparameters separately for student only and the combined student and teacher training. The results of the grid search for initial learning rate $\mu(0)$ and weight decay $\gamma$ for both students on Cityscapes are presented in Table \ref{tab:cs_student_gs}, right of it in Table \ref{tab:student_optimal_hps} are the optimal hyperparameters for all datasets.

%
\begin{table}
    \centering
    \small
    \caption{Grid search results for student-only training. a) shows the whole grid for Cityscapes, b) the best hyperparameters for all datasets. The best performance in a) is highlighted in bold.}
    \begin{subtable}[t]{0.5\hsize}
        \centering
        \caption{Cityscapes Grid Search}
        \label{tab:cs_student_gs}
        \begin{tabular}{lcccc|ccr}\toprule
        &\multicolumn{7}{c}{$\mu(0)$}\\\cmidrule(lr){2-8}
        & \multicolumn{4}{c}{EffNet} & \multicolumn{3}{c}{ResNet}
        \\\cmidrule(lr){2-5}\cmidrule(lr){6-8}
        $\gamma$ & 1e-1 & 5e-2 & 1e-2 & 5e-3 &5e-2 & 1e-2 & 5e-3
        \\\midrule		
            5e-4 & 35.0 & 43.9 & 59.8  & 60.1           &  68.89 & \textbf{70.55} & 67.95   \\
            5e-5 & 62.2 & 63.6  & 61.1 & 60.1            &  69.74 & 69.48  & 67.98           \\
            5e-6 & \textbf{64.6} & 63.6  & 61.1 & 59.4 &  \textbf{70.55} & 67.65 & 67.09 \\\bottomrule
        \end{tabular}
    \end{subtable}
    \begin{subtable}[t]{0.45\hsize}
        \centering
        \caption{All Datasets}
        \label{tab:student_optimal_hps}
        \begin{tabular}{l|cc|cc}\toprule
            & \multicolumn{2}{c}{EffNet} & \multicolumn{2}{c}{ResNet}\\
             & $\mu(0)$ & $\gamma$ & $\mu(0)$ & $\gamma$  \\
            \midrule
            PascalVOC   	& 1e-2 & 5e-4 & 5e-3 & 5e-4 \\
            Cityscapes      & 1e-1 & 5e-6 & 1e-2 & 5e-4  \\
            ADE20K          & 5e-3 & 5e-5 & 1e-2 & 5e-5 \\\bottomrule
        \end{tabular}
    \end{subtable}
\end{table}
We use the same grid for the student and teacher training and set the temperature parameter to $\tau=1$. The results are shown in Tables \ref{tab:teacher_gs} where again the left shows the detailed grid for Cityscapes and the right the optimal hyperparameters for all students and datasets.  

%
\begin{table}
    \centering
    \small
    \caption{Grid search results for student and teacher training. a) shows the whole grid for Cityscapes, b) the best hyperparameters for all datasets. The best performance in a) is highlighted in bold.}
    \label{tab:teacher_gs}
    \begin{subtable}[t]{0.5\hsize}
        \centering
        \caption{Cityscapes Grid Search}
        \label{tab:cs_teacher_gs}
        \begin{tabular}{lcccc|ccr}\toprule
        &\multicolumn{7}{c}{$\mu(0)$}\\\cmidrule(lr){2-8}
        & \multicolumn{4}{c}{EffNet} & \multicolumn{3}{c}{ResNet}
        \\\cmidrule(lr){2-5}\cmidrule(lr){6-8}
        $\gamma$ & 1e-1 & 5e-2 & 1e-2 & 5e-3 &5e-2 & 1e-2 & 5e-3
        \\\midrule		
            5e-4 &  40.4 &  46.9            & 62.8          & 61.1 & 66.44 &	\textbf{71.64} &  70.54 \\
            5e-5 &  62.1 &  65.6            & 63.2          & 62.0 & 67.19 &	70.82 &  68.42 \\
            5e-6 &  65.0 &  \textbf{66.1}   & 63.1          & 61.6 & 70.02 &	69.72 &  68.64 \\\bottomrule
        \end{tabular}
    \end{subtable}
    \begin{subtable}[t]{0.45\hsize}
        \centering
        \caption{All Datasets}
        \label{tab:teacher_optimal_hps}
        \begin{tabular}{l|cc | cc}\toprule
        & \multicolumn{2}{c}{EffNet} & \multicolumn{2}{c}{ResNet}\\
         & $\mu(0)$ & $\gamma$ & $\mu(0)$ & $\gamma$  \\
        \midrule
        PascalVOC & 1e-2 & 5e-6 & 5e-3 & 5e-5 \\
        Cityscapes & 5e-2 & 5e-6 & 1e-2 & 5e-4 \\
        ADE20K     & 1e-2 & 5e-5 & 1e-2 & 5e-5 \\\bottomrule
        \end{tabular}
    \end{subtable}
\end{table}

%
In a second stage of hyperparameter optimization, we tune the temperature parameter $\tau$. The tested values and their responses for all models and datasets can be found in Table \ref{tab:TemperatureImpact}.
A choice of $\tau=1$ appears to work well on PascalVOC and ADE20K, but greater values yield performance gains on Cityscapes. 
\begin{table*}[ht!]
    \centering
    \caption{Impact of the temperature parameter $\tau$ on segmentation performance. The best performances are highlighted in bold. Results within one standard deviation of the best result are underlined. For ADE20K only one run was computed.}
    \label{tab:TemperatureImpact}
    \begin{tabular}{l|cc|cc|cc}\toprule
    &\multicolumn{2}{c}{PascalVOC} & \multicolumn{2}{c}{Cityscapes} & \multicolumn{2}{c}{ADE20K}\\
    $\tau$ &\textbf{EffNet} &\textbf{ResNet} &\textbf{EffNet} &\textbf{ResNet} &\textbf{EffNet} &\textbf{ResNet}\\\midrule
    1   & \textbf{66.24 $\pm$ 0.46}                         & \textbf{65.22 $\pm$ 0.44}         & 65.26 $\pm$ 0.59            & 71.33 $\pm$ 0.85                  &\textbf{36.32}  & \textbf{37.74}\\
    2   & \underline{66.08 $\pm$ 0.16}                      & 64.37 $\pm$ 0.08                  & 64.62 $\pm$ 0.77            & 72.15 $\pm$ 0.10                    & 35.51 & 36.85\\
    3   & \underline{65.91 $\pm$ 0.26}                      & 63.30 $\pm$ 0.44                  & 65.08 $\pm$ 0.63            & \textbf{72.75} $\pm$ \textbf{0.29}  & 34.87 & 36.04\\
    4   &            65.02 $\pm$ 0.82                       & 63.42 $\pm$ 0.54                  & 65.26 $\pm$ 0.43            & \underline{72.44 $\pm$ 0.31}      & 35.02 & 36.34\\
    6   & \underline{66.11 $\pm$ 0.67}                      & 63.65 $\pm$ 0.51                  & 64.52 $\pm$ 0.45            &  \underline{72.49 $\pm$ 0.43}     & 35.23 & 36.49\\
    8   & \underline{66.11 $\pm$ 0.13}                      & 64.18 $\pm$ 0.36                  & \textbf{65.58 $\pm$ 0.03}   & 72.09 $\pm$ 0.50                  & 34.91 & 36.51 
    \\\bottomrule
    \end{tabular}
\end{table*}
%
\subsection{Additional Losses}\label{sec:tuning_losses}
The Losses $L_{PA}$ \cite{Structured}, $L_{HO}$ \cite{Structured}, and $L_{IFV}$ \cite{intra-class} are part of many KD algorithms, but their individual contribution to the student's learning behavior when hyperparameters are optimal is unclear in literature.
We use the optimal hyperparameters of the combined training in Table \ref{tab:teacher_optimal_hps} and optimize the weight of each individual loss term $\{\ \lambda_{PA}, \lambda_{HO}, \lambda_{IFV} \}$ to maximize performance.
\begin{table}
    \small
    \centering
    \caption{Impact of individual loss terms on segmentation performance 
    on Cityscapes. The results
    for three loss terms, $L_{PA}$, $L_{HO}$ and $L_{IFV}$ are presented in tables a), b) and c), respectively. The best 
    \ac{miou} is highlighted in bold for each student model.}
    \label{tab:CS:AdvancedLosses}
    \begin{subtable}[t]{0.3\hsize}
        \centering
        \caption{$L_{PA}$}
        \label{tab:CS:CE_PI_PA}
        \begin{tabular}{lcc}\\\toprule
        & \textbf{EffNet} & \textbf{ResNet}
        \\\cmidrule(lr){2-2}\cmidrule(lr){3-3}
        $\lambda_{PA}$\\\midrule	
        1e-3 & 65.08 & 70.51  \\
        1e-2 & \textbf{65.17} & 70.28  \\
        1e-1 & 63.09 & \textbf{72.98}  \\
        1e+0 & 13.49 & 71.88  \\
        1e+1 & 3.211 & 69.34  \\\bottomrule\\
        \end{tabular}
    \end{subtable}
    \begin{subtable}[t]{0.3\hsize}
    \centering
        \caption{$L_{HO}$}
        \label{tab:CS:CE_PI_HO}
        \begin{tabular}{lcc}\\\toprule
        & \textbf{EffNet} & \textbf{ResNet}
        \\\cmidrule(lr){2-2}\cmidrule(lr){3-3}
        $\lambda_{HO}$\\\midrule	
        1e-4  & \textbf{65.84} & 70.35 \\
        1e-3  & 65.48 & 70.81 \\
        1e-2  & 64.70 & \textbf{72.50} \\
        1e-1  & 41.53 & 28.20 \\\bottomrule
        \end{tabular}
    \end{subtable}
    \begin{subtable}[t]{0.3\hsize}
        \centering
        \caption{$L_{IFV}$}
        \label{tab:CS:CE_PI_IFV}
        \begin{tabular}{lcc}\\\toprule
        & \textbf{EffNet} & \textbf{ResNet}
        \\\cmidrule(lr){2-2}\cmidrule(lr){3-3}
        $\lambda_{IFV}$\\\midrule	
        1e-4  & 65.78 & 70.25  \\
        1e-3  & \textbf{66.01} & 71.70  \\
        1e-2  & 65.80 & \textbf{71.75}  \\
        1e-1  & 64.31 & 71.48  \\
        1e+0  & 65.03 & \textbf{71.75}  \\
        1e+1  & 65.52 & 70.60  \\
        5e+1  & 64.76 & 71.21 \\
        1e+2  & 62.70 & 70.20 \\\bottomrule
        \end{tabular}
    \end{subtable}
\end{table}
The results for the Cityscapes dataset are presented in Table \ref{tab:CS:AdvancedLosses}.


For the EffNet student the optimal values of loss weights are consistently smaller than for the ResNet student. 
\subsection{Final Performance Comparison}
Table \ref{tab:final_comparison} shows the results of the additional loss experiments in comparison to student-only or simple pixel-wise distillation training. It is clear that adding the teacher model and the pixel-wise distillation loss, $L_{PI}$, improves student performance, while none of the three tested additional loss terms provides a further lift. An exception is the PascalVOC dataset, where the conclusion is less clear.  
%
\begin{table*}[ht!]
    \centering
    \caption{Evaluation of more complex loss terms $L_{PA}$, $L_{HO}$, and $L_{IFV}$ when added to $L_{CE}+L_{PI}$ with optimal hyperparameters. Best results are highlighted in bold; results within one standard deviation of the best result are underlined.}
    \label{tab:final_comparison}
    \begin{tabular}{c|c|c|c||cc|cc|cc}
        \toprule
        \multicolumn{4}{c}{} & \multicolumn{2}{c}{Pascal Voc} & \multicolumn{2}{c}{Cityscapes} & \multicolumn{2}{c}{ADE20K}\\
        \textbf{\footnotesize $L_{PI}$} & \textbf{\footnotesize $L_{PA}$} & \textbf{\footnotesize $L_{HO}$} & \textbf{\footnotesize $L_{IFV}$} & \textbf{EffNet} & \textbf{ResNet}  & \textbf{EffNet} & \textbf{ResNet}  & \textbf{EffNet} & \textbf{ResNet} \\\hline
        & & &               & 65.43 $\pm$ 0.38                  & 64.50 $\pm$ 0.34               & 64.42 $\pm$ 0.46             & 70.45 $\pm$ 0.10    & 34.10 & 35.23     \\
        x & & &             & 66.24 $\pm$ 0.46                  & 65.22 $\pm$ 0.44              &  \textbf{65.58 $\pm$ 0.03}    & \textbf{72.75 $\pm$ 0.29}    & \textbf{36.32} &  \textbf{37.74}    \\
        x & x & &           & 66.40 $\pm$ 0.35                  & \textbf{65.84 $\pm$ 0.35}     & 64.94 $\pm$ 0.28              & 71.77 $\pm$ 1.05    & 34.98 & 36.57  \\
        x &     & x &       & \textbf{66.72} $\pm$ \textbf{0.21}& 65.21 $\pm$ 0.28              & 65.00 $\pm$ 0.73              & 71.87 $\pm$ 1.14    & 35.22 &  36.50    \\
        x &     &   & x     & 66.38 $\pm$ 0.23                  & \underline{65.83 $\pm$ 0.39}  & 65.24 $\pm$ 0.70              & 70.96 $\pm$ 1.03    & 35.41 &  37.09   \\\bottomrule
    \end{tabular}
\end{table*}
%

Comparing our results to the related work in Table \ref{tab:over} shows that our student model with a performance of 37.74 clearly outperforms six out of eight methods on ADE20K, even though it was trained using only the most basic distillation loss. As the ADE20K dataset is the most complex dataset with 150 classes, this observation is surprising.
\section{Ablation Study}
Another hyperparameter of KD is the initialization of student weights. The effect of pre-training student backbone models on the ImageNet dataset compared to random weight initialization has been studied with varying conclusions. The authors of SKD find KD to be more efficient when the student is initialized randomly \cite{Structured}. 
CWD \cite{channel-wise} contradicts this by stating that the pre-trained weights help distillation, but calls the lift in student performance less significant compared to the randomly initialized case. 
The provided reason for this statement is that the relative improvement of the student is smaller since the model was better when trained without a teacher. 
\cite{cirkd} and \cite{feat-augment} also report a higher absolute improvement for random initialization. 
\cite{perspective} is very unclear on the subject of weight initialization, stating in a sketch of their algorithm that student weights are initialized randomly, while the performance of student models trained without a teacher suggests an initialization with pre-trained weights.

Our investigation of student weight initialization suggests that distillation of teacher knowledge by the simple pixel-wise distillation loss does not improve the performance of a randomly initialized student, while it does improve the pre-trained one.
\begin{table}[h]
    \centering
    \caption{mIoU of randomly initialized ResNet student trained on Cityscapes compared to literature. The best \ac{miou} is highlighted in bold.
    }
    \label{tab:nopre}
    \begin{tabular}{lccc}\toprule
        Method & SKD & CWD & Ours\\\midrule
        Teacher &  78.56 & 78.5 & 78.24\\\midrule
        $L_{CE}$ & 57.50 & 63.63 & \textbf{63.68 }\\
        $L_{CE}+L_{PI}$  & 58.63 & - & 63.34\\
        $L_{CE}+L_{PI}+L_{PA}+L_{HO}$ & 63.24 & 63.20 & -\\
        \bottomrule
    \end{tabular}
\end{table}
Table \ref{tab:nopre} compares our findings to the results of \cite{Structured} and \cite{channel-wise}. Our experiments find the validation mIoU of the student trained without a teacher to be 63.68, averaged over four runs. This result is in line with CWD, but 6.18 percentage points higher than what is reported by \cite{Structured}. Also similar to CWD we find no positive effect of distillation of teacher knowledge by means of the investigated loss terms.

\section{Environmental Impact}
KD is an effective technique to reduce energy consumption at inference time, as the student model requires less energy than the larger teacher. On the other hand, energy consumption during student training is increased compared to training the student without teacher. These two phenomena result in two trade-offs: the first is performance loss compared to teacher vs. reduced energy consumption during inference and the second is increased performance compared to student-only vs. increased energy consumption during training. The decision about when to use KD always depends on the exact use case. If a model is expected to be deployed on a large number of devices, which all process images at a high rate, energy use during training might be negligible compared to inference and thus KD might be extremely beneficial.  

Using the codecarbon.io tool we calculate the $CO_2$ emissions of training and inference of our experiments with the ResNet student on Cityscapes and show them in Table \ref{tab:co2}. 

\begin{table}
    \centering
    \caption{$CO_2$ emissions of different model combinations at training and inference on Cityscapes. S is the ResNet student, numbers for inference are per 10,000 images.}
    \label{tab:co2}
    \begin{tabular}{lccccc}\toprule
        & & \multicolumn{2}{c}{training} & \multicolumn{2}{c}{inference}
        \\\cmidrule(lr){3-4}\cmidrule(lr){5-6}
        & mIoU & energy (kWh) & $CO_2$ (kg) & energy (kWh) & $CO_2$ (kg) 
        \\\midrule
        S & 70.45 & 0.841 & 0.298 & 0.161 & 0.057 \\
        S + KD & 72.75 & 2.52 & 0.891 & 0.161 & 0.057 \\
        T & 78.24 & 3.01 & 1.06 & 0.413 & 0.146\\
        \\\bottomrule 
    \end{tabular}
\end{table}

Training the student on its own takes 0.841kWh, which with the German electricity conditions means 298g of emitted $CO_2$. As expected, the addition of the teacher in KD leaves energy consumption during inference unchanged, but increases energy consumption and emissions during training to almost the three-fold. Comparing student and teacher reveals that the ResNet student emits only 57g of $CO_2$ per 10,000 images at inference compared to 146g for the teacher model. This means, if we accept the decrease in performance we could save over 60\% of $CO_2$ emission during inference time. 
\section{Conclusion}
In this work, we point out a significant comparability problem in the field of KD for semantic segmentation, which will grow in relevance as more techniques are published. We argue that comparibility can be improved by thoroughly optimizing training hyperparameters and show that doing so eliminates the gains from two accepted techniques, SKD and IFVD, on the two more complex of the three investigated datasets. To facilitate an easier comparison in the future we provide a detailed training protocol including optimal hyperparameters for two student models and three datasets. As part of the hyperparameter tuning we investigate the temperature parameter $\tau$, which most publications in the field simply set to 1. We investigate the entropy of class probability distributions of teacher output to visualize the softening effect of $\tau>1$ and show that the temperature parameter can be beneficial to the distillation of teacher knowledge.  

Following up on this work and using the presented training with optimized hyperparameters, it would be useful to provide a fair comparison of the other loss terms in Table \ref{tab:over}. 
Additionally, the understanding of KD in segmentation could be deepened by an analysis of factors that decide when logit scaling is helpful and why it improves distillation on Cityscapes and not on the other datasets.  


 \section*{Disclaimer}
The results, opinions and conclusions expressed in this publication are not necessarily those of Volkswagen Aktiengesellschaft.

%
%
%
%
%
\bibliographystyle{splncs04}
\bibliography{egbib}

\end{document}